\pdfoutput=1

\documentclass[11pt]{article}

\usepackage[]{acl}

\usepackage{times}
\usepackage{latexsym}

\usepackage{url}
\usepackage{multirow}
\usepackage{graphicx} 
\usepackage{amsmath}
\usepackage{caption,subcaption,booktabs}

\usepackage[T1]{fontenc}

\usepackage[utf8]{inputenc}

\usepackage{microtype}

\title{Easy Adaptation to Mitigate Gender Bias in Multilingual Text Classification}

\author{Xiaolei Huang \\
  Department of Computer Science, University of Memphis \\
  \texttt{xiaolei.huang@memphis.edu} \\
}

\begin{document}
\maketitle
\begin{abstract}
Existing approaches to mitigate demographic biases evaluate on monolingual data, however, multilingual data has not been examined. In this work, we treat the gender as domains (e.g., male vs. female) and present a standard domain adaptation model to reduce the gender bias and improve performance of text classifiers under multilingual settings. We evaluate our approach on two text classification tasks, hate speech detection and rating prediction, and demonstrate the effectiveness of our approach with three fair-aware baselines.
\end{abstract}

\section{Introduction}
Recent research raises concerns that document classification models can be discriminatory and can perpetuate human biases~\cite{dixon2018measuring, borkan2019nuanced, blodgett2020language, liang2020towards}.
Building \textit{fairness}-aware classifiers is critical for the text classification task, such as hate speech detection and online reviews due to its rich demographic diversity of users.
The fairness-aware classifiers aim to provide fair, non-discriminatory outcomes towards people or groups of people based on their demographic attributes, such as gender or race. 
Fairness has been defined in different ways~\cite{hardt2016equality};
for mitigating biases in the text classification, existing research~\cite{dixon2018measuring, heindorf2019debiasing, han2021decoupling} has focused on \textit{group fairness}~\cite{chouldechova2018frontiers}, under which document classifiers are defined as biased if the \textit{classifiers perform better for documents of some groups than for documents of other groups}.

Methods to mitigate demographic biases in text classification task focus on four main directions, data augmentation~\cite{dixon2018measuring, park2018reducing, garg2019counterfactual}, instance weighting~\cite{zhang2020demographics, pruksachatkun2021robustness}, debiased pre-trained embeddings~\cite{zhao2017men, pruksachatkun2021robustness}, and adversarial training~\cite{zhang2018mitigating, barrett2019adversarial, han2021decoupling, liu2021authors}.
The existing studies have been evaluated on English datasets containing rich demographic variations, such as Wikipedia toxicity comments~\cite{cabrera2018gender}, sentiment analysis~\cite{kiritchenko2018examining}, hate speech detection~\cite{huang2020multilingual}.
However, the methods of reducing biases in text classifiers have not been evaluated under multilingual settings.

In this study, we propose a domain adaptation approach using the idea of ``easy adaptation''~\cite{daume2007frustratingly} and evaluate on the text classification task of two multilingual datasets, hate speech detection and rating prediction.
We experiment with non-debiased classifiers and three fair-aware baselines on the gender attribute, due to its wide applications and easily accessible resources.
The evaluation results of both non-debiased and debiased models establish important benchmarks of group fairness on the multilingual settings.
To our best knowledge, this is the first study that proposes the adaptation method and evaluates fair-aware text classifiers on the multilingual settings.

\section{Multilingual Data}

We retrieved two public multilingual datasets that have gender annotations for hate speech classification~\cite{huang2020multilingual} and rating reviews~\cite{hovy2015user}.\footnote{The number of languages is limited by the availability of the data providers.}
The hate speech (\textit{HS}) data collects online tweets from Twitter and covers four languages, including English (en), Italian (it), Portuguese (pt), and Spanish (es). 
The rating review (\textit{Review}) data collects user reviews from Trustpilot website and covers four languages, including English, French (fr), German (de), and Danish (da).
The HS data is annotated with binary labels indicating whether the tweet is related to hate speech or not.
The Review data has five ratings from 1 to 5. 
To keep consistent, we removed reviews with the rating 3 and encoded the review scores into two discrete categories: score > 3 as positive and < 3 as negative. 
All the data has the same categories for the gender/sex, male and female.
We lowercased all documents and tokenized each document by NLTK~\cite{bird2004nltk}, which supports processing English and the other six languages.

\begin{table}[htp]
\centering
\resizebox{0.451\textwidth}{!}{
    \begin{tabular}{c||ccccc}
    Source & Lang & Docs & Tokens & F-Ratio & L-Ratio \\\hline\hline
    \multirow{4}{*}{HS} & EN & 44,253 & 20.533 & .498 & .355 \\
     & IT & 2,361 & 19.848 & .310 & .235 \\
     & PT & 1,852 & 20.007 & .554 & .222 \\
     & ES & 4,831 & 20.660 & .455 & .357 \\\hline\hline
    \multirow{4}{*}{Review} & EN & 358,219 & 48.553 & .398 & .930 \\
     & FR & 324,358 & 37.102 & .429 & .931 \\
     & DE & 115,367 & 38.224 & .430 & .928 \\
     & DA & 882,080 & 49.829 & .475 & .886
    \end{tabular}
}
\caption{Summary of multilingual Hate Speech (HS) and Review data. F-Ratio and L-Ratio indicate female ratios and positive / hate speech label ratios respectively.}
\label{tab:data}
\end{table}

We summarize the data statistics in Table~\ref{tab:data}.
The HS data is comparatively smaller than the review data, and both datasets have a skewed label distributions.
For example, most of the reviews have positive labels.
Notice that the review data comes from a consumer review website in Denmark, and therefore, Danish reviews are more than the other languages of the review data.
We can find that all documents are short and the HS data from Twitter is comparatively shorter.
For the gender ratio, most of the data has a relatively lower female ratios.

\paragraph{Ethic and Privacy consideration.} We only use the text documents and gender information for evaluation purposes without any other user profile, such as user IDs.
All experimental information has been anonymized before training text classifiers.
Specifically, we hash document IDs and replace any user mentions and URLs by two generic symbols, ``user'' and ``url'', respectively.
To preserve user privacy, we will only release aggregated results presented in this manuscript and will not release the data.
Instead, we will provide experimental code and the public access links of the datasets to replicate the proposed methodology.

\section{Easy Adaptation Framework}

\begin{figure*}[htp]
\centering
\includegraphics[width=0.674\textwidth]{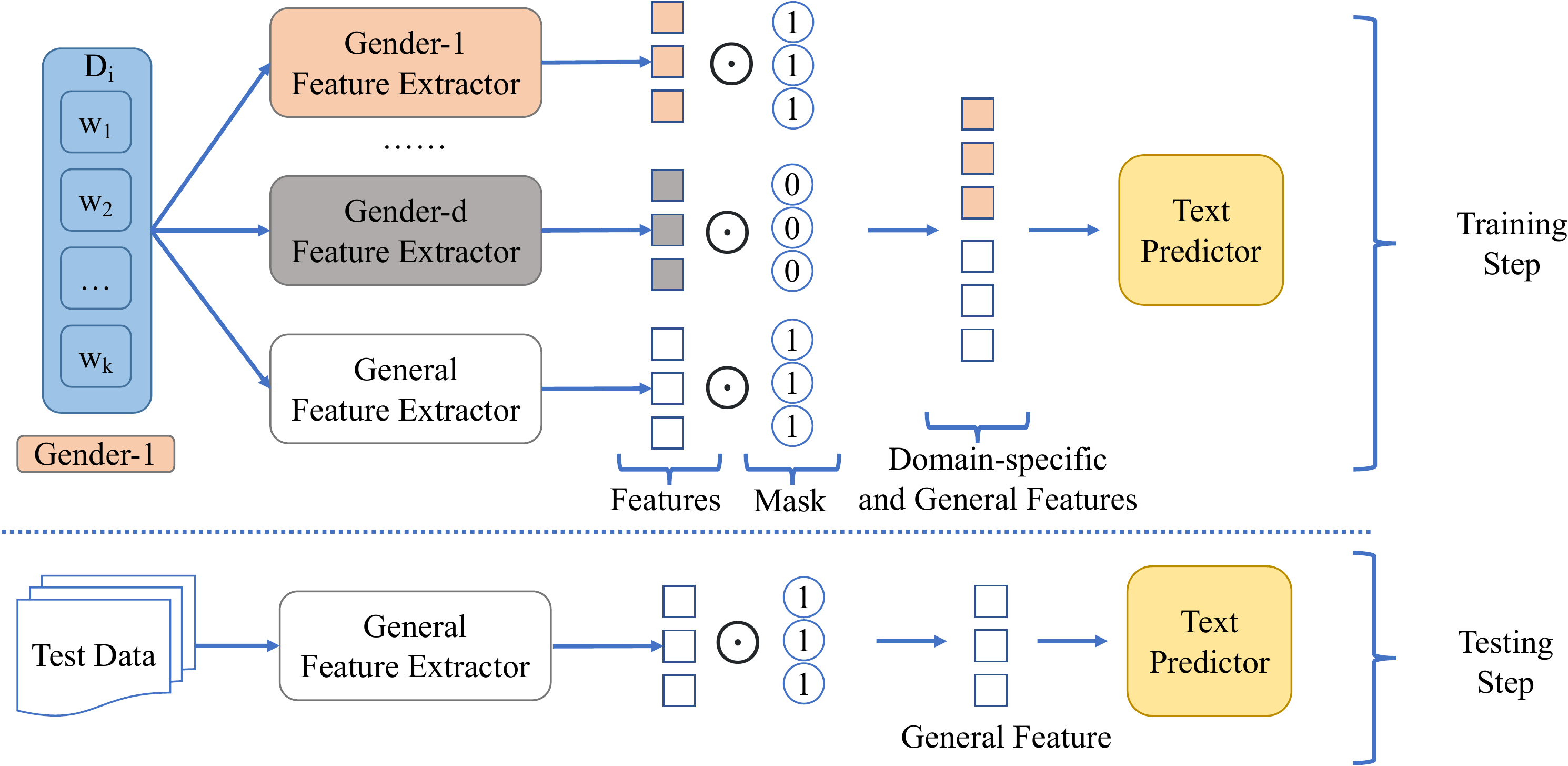}
\caption{Framework illustrations for training and testing steps. The training step uses domain-independent (general in no color) and -specific (in color) features, while the testing step only uses the general features.}
\label{fig:diagram}
\end{figure*}

Previous work has shown that applying domain adaptation techniques, specifically the ``Frustratingly Easy Domain Adaptation'' (\textbf{FEDA}) approach~\cite{daume2007frustratingly},
can improve document classification when demographic groups are treated as domains~\cite{volkova2013exploring, lynn2017human}.
Based on these results, we investigate whether the same technique can also improve the fairness of classifiers, as shown in Figure~\ref{fig:diagram}.
With this method, the feature set is augmented such that each feature has a domain-specific version for each domain, as well as a domain-independent (general) version.
Specifically, the features values are set to the original feature values for the domain-independent features and the domain-specific features that apply to the document, while domain-specific features for documents that do not belong to that domain are set to $0$.
We implement this via a feature mask by the element-wise matrix multiplication.
For example, a training document with a female author would be encoded as $[F_{general}, F_{domain, female}, 0]$, while a document with a male author would be encoded as $[F_{general}, 0, F_{domain, male}]$.
At test time we only use the domain-independent features.
While the FEDA applies to non-neural classifiers, we treat neural models as feature extractors and apply the framework on neural classifiers (e.g., RNN).
We denote models with the easy adaptation with the suffix -DA.

\section{Experiments}
\label{sec:exp}
Demographic variations root in documents, especially in social media data~\cite{volkova2013exploring, hovy2015demographic}.
In this study, we present a standard domain adaptation model on the gender factor, and we treat each demographic group as a domain (e.g., male and female domains).
We show the domain adaptation method can effectively reduce the biases of document classifiers on the two multilingual corpora.
Each corpus is randomly split into training (80\%), development (10\%), and test (10\%) sets. 
We train the models on the training set and find the optimal hyperparameters on the development set. 
We randomly shuffle the training data at the beginning of each training epoch.

\subsection{Regular Baselines (B-Reg)}
We experimented with three popular classifiers, Logistic Regression (LR), Recurrent Neural Network (RNN), and BERT~\cite{devlin2019bert}.
For the LR, we extract Tf-IDF-weighted features for uni-, bi-, and tri-grams on the corpora with the most frequent 15K features with the minimum feature frequency as 3. We then train a \texttt{LogisticRegression} from scikit-learn~\cite{pedregosa2011scikit}. We left other hyperparameters as their defaults.
For the RNN classifier, we follow existing work~\cite{park2018reducing} and build a bi-directional model with the Gated Recurrent Unit (GRU)~\cite{chung2014empirical} as the recurrent unit. 
We set the output dimension of RNN as 200 and apply a dropout on the output with rate .2.
We optimize the RNN with RMSprop~\cite{tieleman2012lecture}.
To encode the multilingual tokens, we utilize the pre-trained fastText multilingual embeddings~\cite{mikolov2018advances} to encode the top 15K frequent tokens.
For the BERT classifier, we build two new linear layers upon on pretrained BERT models~\cite{devlin2019bert} including both English and multilingual versions.
The multilingual version supports 104 languages that cover all languages in this work.
The first layer transforms the BERT-encoded representations into 200-dimension vectors and feeds the vectors for the the final prediction layer.
We optimize the model parameters by the Adam~\cite{kingma2014adam}.
For the neural classifiers, we train them with the batch size as 64, the max length as 200, and the learning rate within the range of $[1e-4, 1e-6]$.
The classifiers in the following sections apply the same hyperparameter settings for fair comparison.

\begin{table*}[ht!]
\centering
\resizebox{.91\textwidth}{!}{
    \begin{tabular}{cc||ccc||ccc||ccc||ccc}
    \multicolumn{2}{c||}{Review (\%)} & \multicolumn{3}{c||}{English} & \multicolumn{3}{c||}{French} & \multicolumn{3}{c||}{German} & \multicolumn{3}{c}{Danish} \\
    \multicolumn{2}{c||}{Methods} & F1-macro & AUC & Fair & F1-macro & AUC & Fair & F1-macro & AUC & Fair & F1-macro & AUC & Fair \\\hline
    \multirow{3}{*}{B-Reg} & LR & 87.1 & 98.3 & 4.2 & 85.1 & 97.9 & 7.7 & 86.1 & 97.9 & 7.6 & 88.5 & 98.4 & 6.2 \\
     & RNN & 87.1 & 97.6 & 5.2 & 80.6 & 95.3 & 1.5 & 80.4 & 95.5 & 3.7 & 86.8 & 94.8 & 3.1 \\
     & BERT & 93.3 & 99.3 & 4.9 & 91.6 & \textbf{99.1} & 4.6 & 91.2 & \textbf{98.2} & 3.5 & \textbf{94.0} & 98.8 & 3.9 \\\hline
    \multirow{5}{*}{B-Fair} & LR-Blind & 87.1 & 98.3 & 3.6 & 85.2 & 97.9 & 7.6 & 86.0 & 97.9 & 5.9 & 90.5 & 98.4 & \textbf{1.4} \\
     & RNN-Blind & 89.6 & 98.5 & 4.5 & 81.7 & 96.0 & 5.1 & 82.0 & 96.8 & 3.2 & 85.8 & 95.7 & 2.5 \\
     & BERT-Blind & 93.4 & 99.3 & 4.3 & \textbf{91.7} & \textbf{99.1} & 3.9 & 89.5 & 98.5 & 3.6 & 93.0 & 99.1 & 1.9 \\
     & RNN-IW & 87.9 & 98.8 & \textbf{2.8} & 81.6 & 97.2 & 4.4 & 84.5 & 97.6 & 3.2 & 86.2 & 97.6 & 1.8 \\
     & RNN-Adv & 88.0 & 98.1 & 5.2 & 83.4 & 96.9 & 4.9 & 85.4 & 97.4 & 3.0 & 88.7 & 97.6 & 1.9 \\\hline
    \multirow{3}{*}{Ours} & LR-DA & 87.3 & 98.4 & \textbf{2.8} & 85.3 & 97.9 & 1.7 & 85.1 & 98.0 & 4.3 & 87.6 & 98.5 & 3.3 \\
     & RNN-DA & 89.2 & 98.6 & 4.1 & 83.1 & 95.9 & \textbf{0.9} & 81.2 & 96.6 & 3.5 & 89.2 & 98.2 & 1.9 \\
     & BERT-DA & \textbf{93.6} & \textbf{99.4} & 3.3 & \textbf{91.7} & 99.0 & 3.4 & \textbf{91.4} & 97.8 & \textbf{2.7} & 93.7 & \textbf{99.2} & 1.7 \\\hline
    \multicolumn{2}{c||}{Delta-R (\%)} & 1.0 & .4 & -28.7 & 1.1 & 0.2 & -56.5 & 0 & 0.3 & -29.1 & 0.5 & 1.3 & -47.7 \\
    \multicolumn{2}{c||}{Delta-F (\%)} & .9 & 0.2 & -16.7 & 2.3 & 0.2 & -61.4 & 0.5 & -0.2 & -7.4 & 1.5 & 1.0 & 21.1\\\hline\hline
    \multicolumn{2}{c||}{Hate Speech (\%)} & \multicolumn{3}{c||}{English} & \multicolumn{3}{c||}{Spanish} & \multicolumn{3}{c||}{Italian} & \multicolumn{3}{c}{Portuguese} \\
    \multicolumn{2}{c||}{Methods} & F1-macro & AUC & Fair & F1-macro & AUC & Fair & F1-macro & AUC & Fair & F1-macro & AUC & Fair \\\hline
    \multirow{3}{*}{B-Regular} & LR & 81.5 & 89.3 & 6.2 & 66.6 & 80.9 & 27.2 & 54.8 & 75.5 & 21.1 & 65.3 & 75.2 & 12.8 \\
     & RNN & 82.0 & 89.0 & 5.4 & 65.3 & 70.0 & 25.9 & 62.3 & 70.7 & 30.9 & 60.8 & 75.9 & 44.1 \\
     & BERT & 84.3 & \textbf{92.0} & 4.9 & 65.9 & 73.8 & 15.6 & 57.1 & 70.3 & 12.9 & 70.1 & 79.6 & 19.9 \\\hline
    \multirow{5}{*}{B-Fair} & LR-Blind & 81.5 & 89.1 & 5.4 & 67.3 & \textbf{81.0} & 25.9 & 54.8 & 75.5 & 20.7 & 62.2 & 73.9 & 9.6 \\
     & RNN-Blind & 82.8 & 89.8 & 5.1 & 64.9 & 63.8 & 14.2 & 56.4 & \textbf{76.4} & 22.9 & 62.2 & 74.9 & 20.6 \\
     & BERT-Blind & 84.0 & 91.9 & 3.7 & 65.5 & 72.8 & 14.9 & 57.2 & 71.2 & 23.2 & 72.4 & \textbf{81.8} & 26.4 \\
     & RNN-IW & 83.8 & 98.4 & 3.8 & 54.0 & 58.9 & 13.4 & 64.1 & 74.7 & 21.9 & 63.8 & 74.7 & 30.7 \\
     & RNN-Adv & 82.9 & 90.6 & 4.1 & 54.6 & 64.8 & 12.0 & 57.9 & 70.9 & 22.1 & 69.8 & 75.8 & 23.1 \\\hline
    \multirow{3}{*}{Ours} & LR-DA & 81.0 & 88.6 & 4.3 & 71.5 & 79.7 & 18.5 & 62.9 & 71.1 & 17.8 & 67.4 & 79.0 & 11.8 \\
     & RNN-DA & 82.1 & 89.1 & 4.7 & 66.5 & 70.9 & 22.8 & 62.8 & 72.3 & 25.6 & 68.8 & 77.1 & 11.7 \\
     & BERT-DA & \textbf{84.4} & 91.4 & \textbf{2.2} & \textbf{73.8} & 78.3 & \textbf{10.1} & \textbf{67.2} & 74.9 & \textbf{12.4} & \textbf{74.8} & 78.3 & \textbf{9.0} \\\hline
    \multicolumn{2}{c||}{Delta-R (\%)} & -0.1 & -0.4 & -32.1 & 7.1 & 1.9 & -25.2 & 10.7 & 0.8 & -14.0 & 7.5 & 1.6 & -57.7 \\
    \multicolumn{2}{c||}{Delta-F (\%)} & -0.6 & -2.5 & -15.5 & 15.2 & 11.8 & 6.6 & 10.7 & -1.3 & -16.1 & 6.4 & 2.5 & -50.9
    \end{tabular}
}

\caption{Performance on the HS and Review Data in percentage. A lower fair score is better. The Delta-R and -F are improvements over the regular (-R) and fair (-F) baselines respectively. Negative Delta scores over the fair indicate percentage of mitigating biases, and lower scores means more bias mitigation.}
\label{tab:results}
\end{table*}

\subsection{Fair-aware Baselines}
\paragraph{Blind} augments data by masking out tokens that are associated with the demographic groups~\cite{dixon2018measuring, garg2019counterfactual}.
We apply the Blind strategy on the regular baselines and denote the classifiers as LR-Blind, RNN-Blind, and BERT-Blind respectively.
We retrieved the gender-sensitive tokens from the Conversation AI project~\cite{conversationai2021}, which contains individual tokens.
However, the existing resource~\cite{dixon2018measuring,garg2019counterfactual} only focused on English instead of the other languages.
Therefore, we use the multilingual lexicon, PanLex~\cite{kamholz2014panlex}, to translate the gender-sensitive English tokens into the other six languages.

\paragraph{RNN-IW} applies the instance weighting to reduce impacts of gender-biased documents~\cite{zhang2020demographics} during training classifiers.
The method learns each training instance with a numerical weight $\frac{P(Y)}{P(Y|Z)}$ based on explicit biases counted by gender-sensitive tokens~\cite{conversationai2021}, utilizes a random forest classifier to estimate the conditional distribution $P(Y|Z)$ and the marginal distribution $P(Y)$, applies the classifier on training instances to obtain training weights.
The approach achieves the best results using RNN models, and we keep the same settings.
We extend the approach to multilingual settings using the translated resources.

\paragraph{RNN-Adv} utilizes adversarial training~\cite{han2021decoupling} to mitigate~\cite{liu2021authors} gender biases by two prediction tasks, document and gender predictions.
Instead of better separating document labels, the adversarial training aims to confuse the gender predictions to reduce gender sensitiveness.
We adapt the RNN module which achieved promising results~\cite{han2021decoupling, liu2021authors}.

\subsection{Evaluation Metrics}
We use F1-macro score (fit for skewed label distribution) and area under the ROC curve (AUC) to measure overall performance.
To evaluate {group fairness},
we measure the \textit{equality differences} (ED) of false positive/negative rates~\cite{dixon2018measuring} for the fair evaluation.
Existing study shows the FP-/FN-ED is an ideal choice to evaluate fairness in classification tasks~\cite{czarnowska2021quantifying}.
Taking the false positive rate (FPR) as an example, we calculate the equality difference by $FPED = \sum_{g \in G}|FPR_d - FPR|$, where $G$ is the gender and $d$ is a gender group (e.g., female).
We report the sum of FP-/FN-ED scores and denote as ``Fair''.
This metric sums the differences between the rates within specific gender groups and the overall rates.

\subsection{Results}
We present the averaged results after running evaluations three times of both baselines and our approach in Table~\ref{tab:results}.
Fair-aware classifiers have significantly reduced the gender bias over regular classifiers across the multilingual datasets, and our approaches have better scores of the group fairness by a range of 14\% to 57.7\% improvements over the baselines.
The data augmentation approach achieves better fair scores across multiple languages, which indicates that the translated resources of English gender-sensitive tokens can also be effective on the evaluated languages.
The neural fair-aware RNNs usually achieve worse performance than the BERT-based models.
Note that the BERT and fastText embeddings were pretrained on the same text corpus, Wikipedia dumps, and the performance indicates that fine-tuning the more complex models is a practical approach to reduce gender bias under the multilingual settings.
Overall, our approach appears promising to reduce the gender bias under the multilingual setting.

Considering model performance, we can generally find that the fair-aware methods do not significantly improve the model performance, which aligns with findings in a previous study~\cite{menon2018cost}.
However, we also find that all fair-aware models achieve better performance on the Spanish, Italian, and Portuguese hate speech data.
We infer this due to the data size, as for the three data are much smaller than the other corpora.

\section{Conclusion}
We present an easy adaptation method to reduce gender bias under the multilingual setting.
The experiments show that by treating demographic groups as domains, we can reduce biases while keeping relatively good performance.
Our future work will solve the limitations of this study, including non-binary genders, multiple demographic factors, data sizes, and label imbalance.
Code and data instructions of our work are available at \url{https://github.com/xiaoleihuang/DomainFairness}.

\section*{Acknowledgements}
The author wants to thank for reviewers' valuable comments. The experiments on the English hate speech data was adopted from his Ph.D. thesis, Section 5.6~\cite{huang2020metadata}. The author also wants to thank Dr. Michael J. Paul for the initial idea discussion.

\bibliography{custom}

\appendix

\section{Implementation Details}

While we have presented experimental and hyperparameter settings in the Section~\ref{sec:exp}, we report implementation tools in this section.
We implement neural models by PyTorch~\cite{adam2019pytorch} and non-neural models by scikit-learn~\cite{pedregosa2011scikit}. %
For the BERT model, we use the Hugging Face Transformers~\cite{wolf2020transformers}.
The Keras~\cite{chollet2015keras} helped preprocess text documents for neural models, including padding and tokenization.
We trained models on an NVIDIA RTX 3090 and evaluated the models on CPUs.

\section{Limitations and Future work}

While we have proved the effectiveness of our proposed framework, limitations must be acknowledged in order to appropriately interpret our evaluations. Our experiments are based on coarse-grained gender categories (binary gender groups) and the multilingual datasets fail to provide fine-grained information.
Using coarse-grained attributes would ignore people with non-binary gender.
Expanding evaluations of existing methods may require enriching categories of demographic attributes.
In this study, we include two major data sources and experiment the six languages aiming to evaluate gender-bias-mitigation algorithms in a diverse and multilingual scenario.
We keep the same experimental settings with the baselines~\cite{dixon2018measuring, zhang2020demographics, han2021decoupling, liu2021authors} to ensure fair comparisons, such as data sources and binary labels.



\end{document}